\documentclass[letterpaper, 10 pt, conference]{ieeeconf}

\IEEEoverridecommandlockouts

\title{\LARGE \bf
Uncertainty-Driven Efficient Communication for UAV-Based Collaborative Perception
}

\author{Jiuwu Hao$^{1,*}$, Yuting Wan$^{1,*}$, Liguo Sun$^{2}$, Zao Zhang$^{2}$, Yueyang Wu$^{1}$ and Pin Lv$^{2,\dagger}$%
\thanks{$^{1}$Jiuwu Hao, Yuting Wan and Yueyang Wu are with School of Artificial Intelligence, University of Chinese Academy of Sciences, and with Institute of Automation, Chinese Academy of Sciences
        }%
\thanks{$^{2}$Liguo Sun, Zao Zhang and Pin Lv are with Institute of Automation, Chinese Academy of Sciences
        }%
\thanks{$^{*}$~Equal contribution}        
\thanks{$^{\dagger}$~Corresponding author}}%

\usepackage{cite}
\usepackage{amsmath}
\usepackage{amssymb}
\usepackage{amsfonts}
\usepackage{bbm}
\usepackage{graphicx}
\usepackage{subcaption}
\usepackage{balance}
\usepackage{amssymb}
\usepackage{pifont}

\usepackage{array}
\usepackage{color, xcolor}
\usepackage{colortbl}
\usepackage{booktabs}
\usepackage{multicol}
\usepackage{multirow}
\usepackage{adjustbox}

\newcommand{\modelname}{\texttt{UCP}}

\usepackage[colorlinks=true, linkcolor=blue, citecolor=blue, urlcolor=blue]{hyperref}

\begin{document}

\maketitle
\thispagestyle{empty}
\pagestyle{empty}

\begin{abstract}

Unmanned aerial vehicle (UAV)-based collaborative perception enables multiple agents to share complementary perceptual information, overcoming limitations such as occlusion and limited sensing range. Communication efficiency is a critical bottleneck, as UAVs operate under stringent bandwidth and computational constraints. However, existing spatial selection methods focus predominantly on foreground regions to reduce missed detections, neglecting perceptually ambiguous regions and leaving false detections unresolved. We propose \texttt{UCP}, an uncertainty-driven framework that characterizes collaborative value through observational uncertainty, identifying sensor-limited regions, and discriminative uncertainty, capturing model-level ambiguity near the decision boundary. These estimates drive spatial request and supply maps for efficient inter-agent communication, and the received messages are aggregated via U-Fusion, an uncertainty-guided cross-attention module. Experiments on three benchmarks across 3D object detection and trajectory prediction demonstrate a superior performance-bandwidth trade-off with high interpretability and strong generalizability.

\end{abstract}
\section{Introduction}

UAV-based collaborative perception has emerged as a promising paradigm that enables multiple agents to share complementary perceptual information, effectively overcoming the inherent limitations of single-agent systems such as occlusion and limited sensing range~\cite{UAV3D, DHD, AGC-Drive}. This paradigm has demonstrated significant potential across a broad range of applications, including traffic monitoring, search and rescue, and autonomous driving support.

Despite these advances, communication efficiency remains a critical bottleneck in practice, as UAVs are subject to stringent constraints on bandwidth, computational resources, and energy consumption.
Among existing approaches, spatial selection mechanisms based on Bird's Eye View (BEV) representations have emerged as a particularly promising direction, owing to their efficiency and high interpretability. However, existing methods typically focus on sparse background regions to address missed detections through collaboration~\cite{Where2comm, CoSDH}, while overlooking perceptually ambiguous regions where the single-agent model struggles to make confident decisions. This asymmetric focus leaves false detections largely unaddressed, limiting the overall effectiveness of collaboration.

We observe that the uncertainty of a perception model is a strong indicator of its failure modes. As shown in Fig.~\ref{fig:figure02}, incorrect predictions consistently exhibit significantly higher uncertainty than correct ones, and correcting high-uncertainty regions yields substantially larger improvements in detection performance. These observations motivate us to go beyond objectness-based spatial selection and additionally consider the uncertainty of each spatial location as a key criterion for determining collaborative priority.

Grounded in this insight, we propose \modelname, a novel uncertainty-driven collaborative perception framework that characterizes the collaborative value of each spatial location through two complementary forms of uncertainty. Each agent first estimates its intra-agent uncertainty, comprising observational uncertainty to identify sensor-limited regions and discriminative uncertainty to capture model-level ambiguity. These uncertainty estimates are then used to construct spatial request and supply maps, which jointly determine where each agent needs assistance and where it can provide reliable support. Guided by these maps, the most valuable spatial regions are selected for transmission, and the received messages are aggregated through an uncertainty-guided cross-attention fusion module (U-Fusion) that explicitly conditions attention on the spatial demand and reliability of each collaborator.

\begin{figure}[!t]
  \centering
  \includegraphics[width=\linewidth, keepaspectratio]{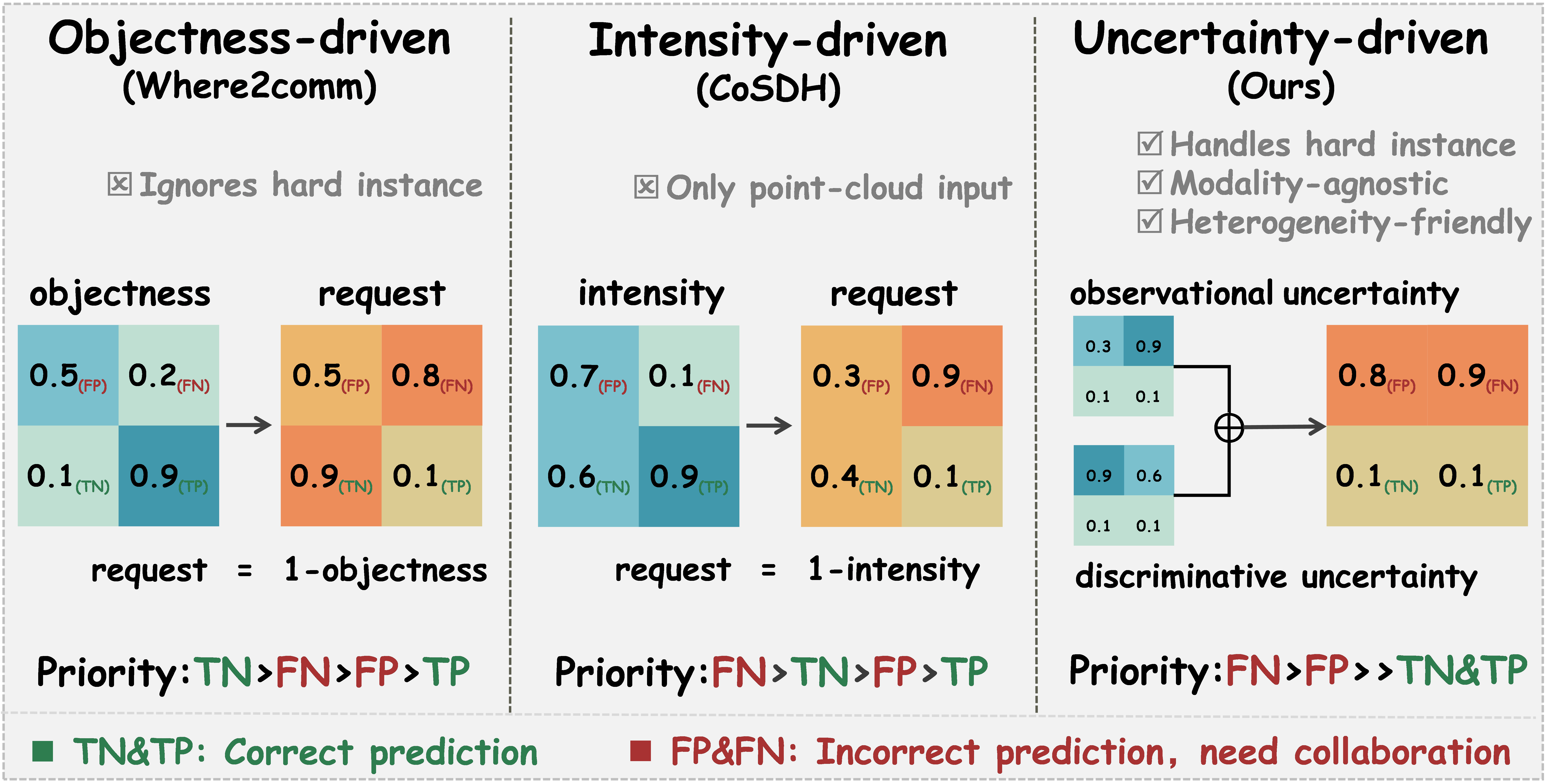}
    \caption{Comparison of spatial selection strategies for collaborative perception. The proposed uncertainty-driven strategy combines observational uncertainty and discriminative uncertainty to jointly identify observation-limited and perceptually ambiguous regions, assigning the highest collaboration priority to incorrectly predicted regions.}
  \label{fig:figure01}
\end{figure}

As illustrated in Fig.~\ref{fig:figure01}, the proposed {\modelname} offers three key advantages over existing methods. First, it is \textit{heterogeneity-friendly}: since the transmitted messages consist of modality-agnostic semantic perception results paired with uncertainty estimates, {\modelname} supports collaboration among agents with heterogeneous sensor configurations without any retraining. Second, it achieves \textit{high interpretability}: the uncertainty maps explicitly distinguish whether a spatial region suffers from observation limitations or model-level ambiguity, providing physically meaningful guidance for spatial selection. Third, {\modelname} achieves a \textit{superior performance-bandwidth trade-off} by concentrating communication on the most uncertain and informative spatial regions, significantly reducing redundant transmission while maintaining high perception quality.

To validate the effectiveness of \texttt{{\modelname}}, we conduct extensive experiments across multiple tasks, including 3D object detection and trajectory prediction, on three UAV-based collaborative perception datasets spanning both simulation and real-world settings. The results consistently demonstrate the effectiveness and robustness of the proposed framework.

The main contributions of this work are as follows:
\begin{itemize}
    \item We propose \modelname, a unified uncertainty-driven collaborative perception framework that jointly models spatial demand and supply through intra-agent uncertainty estimation, enabling efficient and interpretable inter-agent communication.
    \item We introduce a dual uncertainty characterization combining observational uncertainty and discriminative uncertainty, which captures complementary failure modes of single-agent perception and enables a superior performance-bandwidth trade-off compared to existing spatial selection methods.
    \item Extensive experiments on three UAV-based collaborative perception benchmarks across diverse tasks and settings validate the effectiveness and generalizability of the proposed framework.
\end{itemize}

\section{Related Work}

\textbf{Collaborative Perception.} 
Collaborative perception is a promising solution to overcome the inherent limitations of single-agent sensing, enabling agents to share complementary observations and achieve a more comprehensive understanding of the environment. UAV-based collaborative perception has attracted growing attention owing to the wide field of view and operational flexibility that aerial platforms provide. This emerging field has witnessed a surge of dedicated datasets to support algorithmic development~\cite{DHD, U2UData, UAV3D, U2UDataPlus, Griffin, AGC-Drive, AirV2X, AirCopBench, V2U4Real}. Building on these resources, recent works have demonstrated consistent gains across diverse UAV-based perception tasks, including object detection~\cite{UCDNet, C2F-Net}, trajectory forecasting~\cite{DHD}, and occupancy prediction~\cite{MCOP}. In this work, we focus on the trade-off between perception performance and communication overhead, which is essential for the scale-up of UAV-based collaborative perception in real-world deployment.

\begin{figure}[t]
    \centering
    \begin{subfigure}[b]{0.48\columnwidth}
        \centering
        \includegraphics[width=\linewidth]{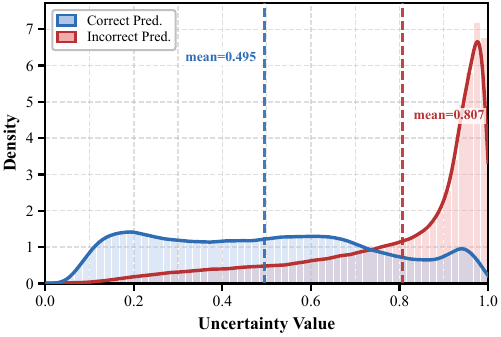}
        \caption{}
        \label{fig:sub1}
    \end{subfigure}
    \hfill
    \begin{subfigure}[b]{0.48\columnwidth}
        \centering
        \includegraphics[width=\linewidth]{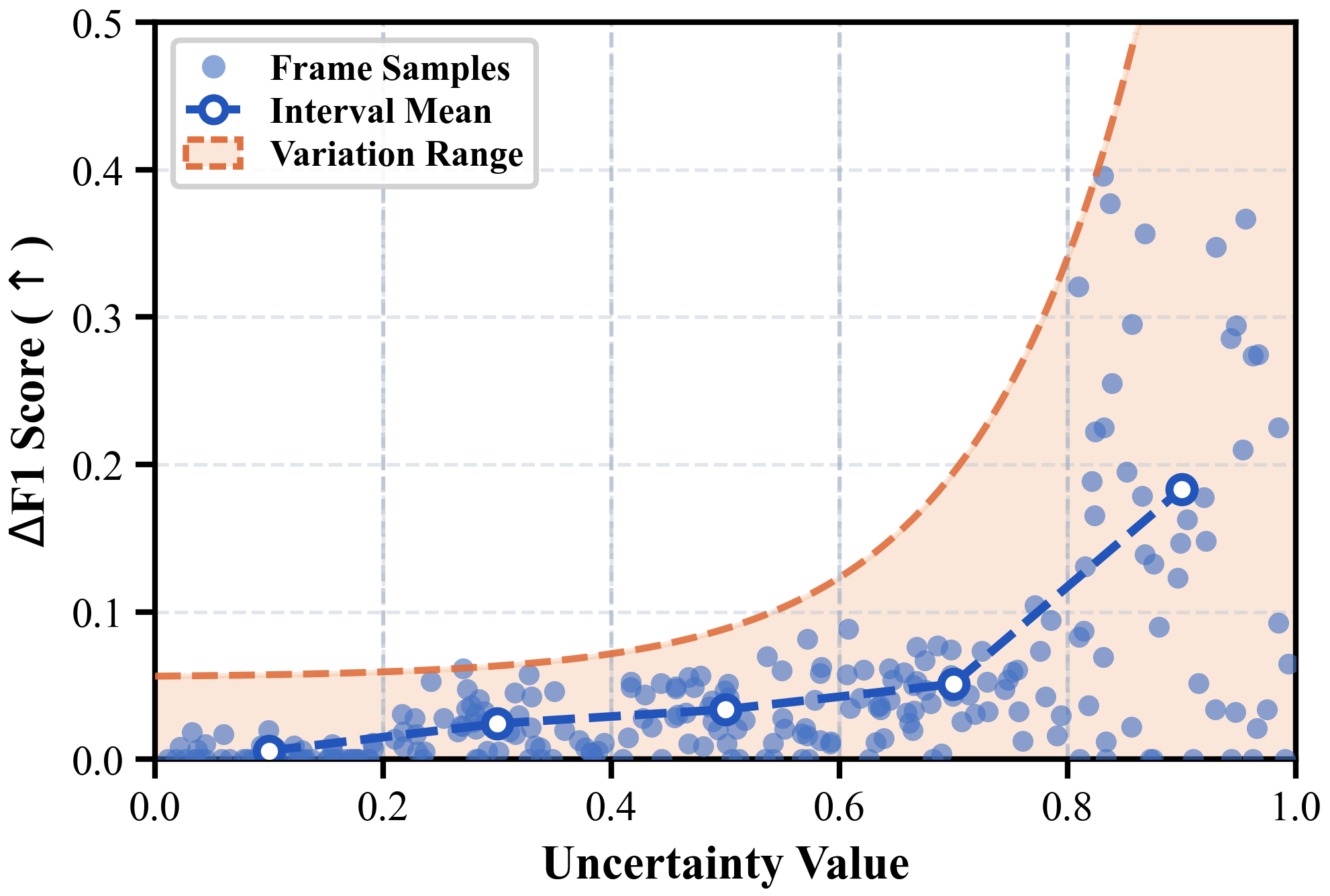}
        \caption{}
        \label{fig:sub2}
    \end{subfigure}
    \caption{Statistical analysis of prediction uncertainty. (a) Incorrect predictions exhibit significantly higher uncertainty than correct ones. (b) Correcting high-uncertainty regions yields larger performance gains, motivating uncertainty-driven spatial selection.}
    \label{fig:figure02}
\end{figure}

\textbf{Spatial selection for efficient communication.}
Considering that communication efficiency is the bottleneck issue in multi-agent collaborative perception, prior works have explored several strategies, including partner identification~\cite{When2com, Who2com}, message representation~\cite{CodeFilling}, spatial selection~\cite{Where2comm, CoSDH, Birdcast, LIF, GevBEV, UMC}, and temporal aggregation~\cite{Pragcomm}. Among them, spatial selection has been widely adopted for its generalizability and interpretability. Where2comm~\cite{Where2comm} is the first to leverage spatial confidence maps to guide the communication process, and has since inspired a line of follow-up works~\cite{Confidence-V2X, EIMC}. CoSDH~\cite{CoSDH} and QTF~\cite{QTF} adopt local point cloud density as an indicator of collaborative necessity. DHD~\cite{DHD} and UMC~\cite{UMC} quantify the collaborative value of each spatial region by measuring local feature discrepancy within a sliding window. Birdcast~\cite{Birdcast} constructs maps of interest by combining local information density with inter-agent confidence gaps. In contrast, we propose to leverage the intrinsic observational and discriminative uncertainties of individual agents to model the spatial supply-demand relationship for collaboration, achieving a superior perception-communication trade-off.

\textbf{Uncertainty estimation.}
Uncertainty estimation has been widely studied as a principled approach to quantifying the reliability of model predictions in perception tasks. GevBEV~\cite{GevBEV} employ a learnable uncertainty estimation framework to identify and prioritize spatially informative regions for transmission. CoAlign~\cite{CoAlign} estimates bounding box uncertainty to identify reliable scene landmarks for pose graph optimization, improving robustness against localization errors. Double-M~\cite{Double-M} captures both epistemic and aleatoric uncertainty of collaborative detection via a moving block bootstrap algorithm, providing principled uncertainty quantification for safety-critical applications. CoCa3D~\cite{CoCa3D} selects depth features with low pixel-wise entropy for sharing and uses detection confidence to identify perceptually critical regions, improving camera-only collaborative 3D detection.

\section{Problem Formulation}
\begin{figure*}[htbp]
  \centering
  \includegraphics[width=\linewidth, keepaspectratio]{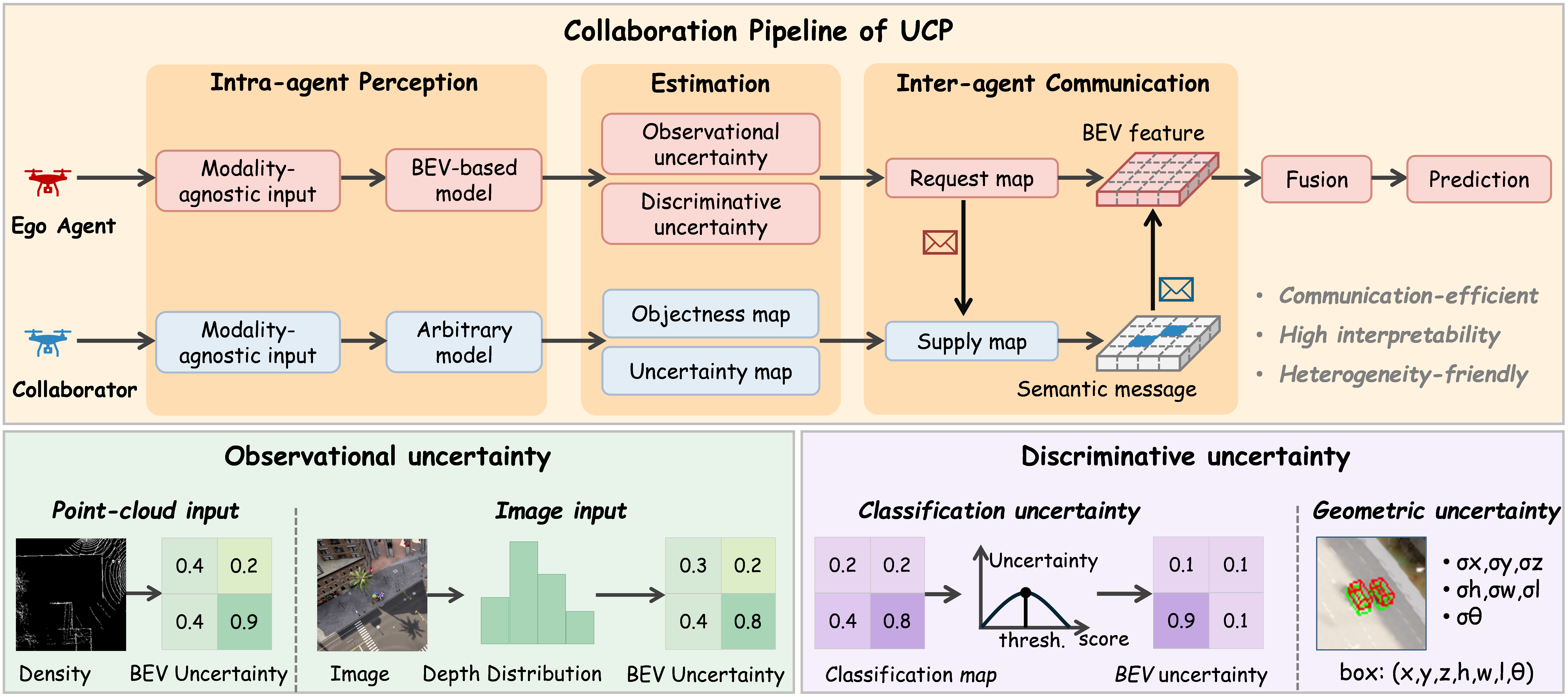}
    \caption{Overview of the proposed {\modelname} framework. Each agent estimates intra-agent uncertainty from single-agent perception, comprising observational uncertainty (derived from point cloud density for LiDAR-based agents or depth estimation confidence for camera-based agents) and discriminative uncertainty (classification uncertainty and geometric uncertainty). These estimates are used to construct a request map for the ego agent and a supply map for the collaborator, which jointly drive spatial selection of the most valuable regions for transmission. The transmitted semantic message is then fused with the ego BEV feature via the proposed U-Fusion, an uncertainty-guided cross-attention module, to produce the final perception result.}
  \label{fig:overview}
\end{figure*}
Consider $N$ agents in the scene. Each agent $i$ is equipped with its own sensor modality and a corresponding perception model. Let $\mathcal{X}_i$ denote the raw observation of agent $i$, and let $\mathcal{Y}_i$ denote the corresponding ground-truth supervision. Each agent concurrently acts as both a supporter and a receiver: as a supporter, it contributes perceptual messages to assist others; as a receiver, it aggregates messages from other agents to enhance its own perception.

Let $P_{j \rightarrow i}$ denote the message transmitted from agent $j$ to agent $i$. The objective of collaborative perception is to maximize the perception performance of all agents under a communication budget $B$:

\begin{equation}
    \max_{\theta, \mathcal{P}} \sum_{i=1}^{N} g\left(\Phi_{\theta}\left(\mathcal{X}_i, \{P_{j \rightarrow i}\}_{j=1, j \neq i}^{N}\right), \mathcal{Y}_i\right),
\end{equation}
\begin{equation}
    \text{s.t.} \quad \sum_{i,j, j \neq i} b(P_{j \rightarrow i}) \leq B,
\end{equation}

where $g(\cdot, \cdot)$ is the evaluation metric for the perception task, $\Phi_{\theta}$ is the collaborative perception model with trainable parameters $\theta$, and $b(\cdot)$ measures the communication cost of the transmitted message.

\section{Method}

\subsection{Overall Architecture}

The overall architecture of the proposed method is illustrated in Fig.~\ref{fig:overview}. Each agent first processes its raw observation $\mathcal{X}_i$ through a modality-specific perception model, such as PointPillars~\cite{PointPillars} for point clouds or BEVFusion~\cite{BEVFusion} for multi-modal inputs, to generate a BEV feature map $\mathcal{F}_i$ and a single-agent perception result $\mathcal{O}_i$. Each agent derives a spatial supply map and a demand map, which characterize where it can provide reliable perceptual information and where it requires assistance from others, respectively. Guided by these maps, each agent selects and transmits compact yet informative messages to its collaborators. For homogeneous agents, BEV features can additionally be included in the transmitted messages to further enhance collaboration. Upon receiving messages from other agents, each agent fuses them with its local BEV feature $\mathcal{F}_i$ to obtain an enhanced representation, which is subsequently passed to the perception head to produce the final collaborative perception result $\hat{\mathcal{O}}_i$.

\subsection{Intra-Agent Uncertainty Estimation}

Single-agent perception failures can be attributed to two distinct sources: i) observation-limited regions, where insufficient sensor coverage leads to missed detections, and ii) perceptually ambiguous regions, where hard instances near the decision boundary induce significantly elevated model uncertainty and degrade detection reliability. These two types of regions represent the highest collaborative value, yet existing works fail to account for both simultaneously. This observation motivates us to characterize collaboration priority through two complementary forms of perceptual uncertainty, namely observational uncertainty and discriminative uncertainty, corresponding to observation-limited regions and perceptually ambiguous regions, respectively.

\subsubsection{Observational Uncertainty Estimation}

Observational uncertainty reflects the degree to which an agent's sensor fails to adequately cover a given spatial region, and serves as the demand signal indicating where collaborative assistance is needed.

For LiDAR-based agents, following~\cite{CoSDH}, we use local point cloud density as the indicator: regions with sparse or absent point cloud returns correspond to distant or occluded areas that are poorly observed, and are thus assigned high observational uncertainty. For agent $i$, the point cloud density at each BEV cell is computed by counting the number of points within the corresponding pillar and normalizing to $[0, 1]$, yielding a density map $A_i \in [0,1]^{H \times W}$. The observational uncertainty map is then defined as:
\begin{equation}
    U^{obs}_i = 1 - A_i.
\end{equation}

For camera-based agents, we propose to derive observational uncertainty from the depth estimation confidence of the LSS~\cite{LSS} depth network. The depth network predicts a per-pixel categorical depth distribution over $D$ discretized depth bins via softmax. A confident depth prediction concentrates probability mass on a single bin, producing a high-density BEV projection at the corresponding spatial location, whereas an uncertain prediction yields a diffuse distribution and low BEV density. Formally, let $p_i^n(h,w) \in \mathbb{R}^D$ denote the depth probability of pixel $(h,w)$ from camera $n$, and let $\pi(n,d,h,w)$ denote its projected BEV coordinate at depth bin $d$. The depth-probability-weighted BEV accumulation is:
\begin{equation}
    \tilde{A}_i(u,v) = \sum_{n,d,h,w} p_i^n(h,w,d) \cdot \mathbf{1}[\pi(n,d,h,w) = (u,v)],
\end{equation}
where $\mathbf{1}[\cdot]$ denotes the indicator function. The accumulation is then normalized to obtain $A_i = \tilde{A}_i / \max(\tilde{A}_i) \in [0,1]^{H \times W}$. The observational uncertainty map is defined as $U^{obs}_i = 1 - A_i$, with high values indicating regions where depth perception is unreliable and collaborative assistance is most beneficial.

\subsubsection{Discriminative Uncertainty Estimation}

Beyond observational uncertainty, we propose discriminative uncertainty to jointly assess the collaborative value of each spatial region, which can be regarded as the systematic uncertainty intrinsic to the single-agent perception model. Taking object detection as the representative task, discriminative uncertainty can be decomposed into two components: classification uncertainty, which reflects ambiguity in determining whether a region contains an object, and geometric uncertainty, which captures imprecision in the geometric estimation of object position and extent.

\textbf{Classification uncertainty.}
Let $s_i(h,w) \in [0,1]$ denote the detection confidence at BEV cell $(h,w)$. A piecewise linear mapping is applied to transform $s_i$ into a probability $p_i$ that reflects its proximity to the decision boundary:
\begin{equation}
    p_i(h,w) = \begin{cases} \dfrac{s_i(h,w)}{2\tau} & s_i(h,w) < \tau, \\[6pt] \dfrac{1}{2} + \dfrac{s_i(h,w) - \tau}{2(1-\tau)} & s_i(h,w) \geq \tau, \end{cases}
\end{equation}
where $\tau$ is the detection threshold. The classification uncertainty map is then defined as the binary entropy of $p_i$:
\begin{equation}
    U^{cls}_i(h,w) = -p_i \log_2 p_i - (1-p_i)\log_2(1-p_i) \in [0,1].
\end{equation}
The map $U^{cls}_i$ reaches its maximum of 1 when $s_i = \tau$, corresponding to the most ambiguous regions, and approaches 0 for confidently detected foreground or clearly observed background.

\textbf{Geometric uncertainty.}
Geometric uncertainty characterizes the imprecision of each detected bounding box in terms of center position, heading angle, and box dimensions. Following~\cite{CoAlign}, a lightweight uncertainty head is attached to the BEV feature map to predict per-location log-uncertainty values. Specifically, for each BEV cell, the uncertainty head outputs five channels: $\log\sigma^2_x$ and $\log\sigma^2_y$ for center position, $\log\kappa_\theta$ for heading angle, and $\log\sigma^2_w$ and $\log\sigma^2_l$ for box dimensions. Center position and box dimensions are modeled as Gaussian distributions, while the heading angle, due to its periodicity, is modeled as a Von-Mises distribution with concentration parameter $\kappa_\theta = e^{-s_\theta}$.

\subsection{Uncertainty-Driven Inter-Agent Communication}

Based on the intra-agent uncertainty estimates, spatial request and supply maps are constructed to model the collaborative demand and contribution of each agent, and the most valuable regions are selected for communication.

\textbf{Request map.}
The request map $R_i \in [0,1]^{H \times W}$ of agent $i$ characterizes where collaborative assistance from other agents is needed. It is defined as a weighted combination of the three uncertainty components derived in the previous section:
\begin{equation}
    R_i = \alpha_1 U^{obs}_i + \alpha_2 U^{cls}_i + \alpha_3 U^{geo}_i,
\end{equation}
where $\alpha_1$, $\alpha_2$, and $\alpha_3$ are hyperparameters. Regions with high request values correspond to areas where the perception of the agent is either observation-limited or model-uncertain, and thus benefit most from external collaborative information.

\textbf{Supply map.}
The supply map $S_j \in [0,1]^{H \times W}$ of agent $j$ characterizes where it can provide reliable and informative perceptual support to others. It is defined as a weighted combination of an objectness score and a certainty score:
\begin{equation}
    S_j = \beta_1 \cdot \mathrm{Obj}_j + \beta_2 \cdot (1 - U_j),
\end{equation}
where $\mathrm{Obj}_j$ is the objectness map, $U_j = \alpha_1 U^{obs}_j + \alpha_2 U^{cls}_j + \alpha_3 U^{geo}_j$ is the aggregated uncertainty of agent $j$, and $\beta_1$, $\beta_2$ are hyperparameters. Regions with high supply values indicate that agent $j$ has confident detections and reliable perception, making them valuable candidates for transmission.

\textbf{Spatial selection and communication.}
Given the request map of agent $i$ and the supply map of agent $j$, the spatial collaboration score is defined as:
\begin{equation}
    \mathcal{M}_{j \rightarrow i}(h,w) = R_i(h,w) \cdot S_j(h,w).
\end{equation}
The top-$K$ spatial locations with the highest collaboration scores are selected for transmission, where $K = \lfloor \rho \cdot H \cdot W \rfloor$ and $\rho$ is the communication ratio. To support heterogeneous agent collaboration, the transmitted message consists of compressed semantic perception results, i.e., detection outputs paired with their associated uncertainty estimates. This representation is modality-agnostic and thus compatible across agents with different sensor configurations. For homogeneous agents, BEV features can additionally be included in the transmitted message to further enhance collaborative fusion. Upon receiving the messages, each agent fuses them with its local BEV representation to produce the final collaborative perception result.

\subsection{Uncertainty-Guided Collaborative Fusion}
Upon receiving messages from collaborating agents, the detection outputs and associated uncertainty estimates are scattered onto the BEV grid to form a sparse semantic feature map $\mathcal{V}_{j \rightarrow i} \in \mathbb{R}^{H \times W \times C_s}$. To adaptively aggregate the received messages, we design an uncertainty-guided cross-attention fusion module, termed \textit{U-Fusion}, in which the query is derived from the ego BEV feature $\mathcal{F}_i$, and the key and value are derived from $\mathcal{V}_{j \rightarrow i}$. The attention score is jointly modulated by the ego request map $R_i$ and the supply map of collaborator $j$:
\begin{equation}
    e_{j\rightarrow i}(h,w) = \frac{\mathbf{Q}(h,w)^\top \mathbf{K}(h,w)}{\sqrt{d}} \cdot R_i(h,w) \cdot S_j(h,w),
\end{equation}
\begin{equation}
    \mathbf{A}_{j\rightarrow i}(h,w) = \mathrm{softmax}\left(e_{j\rightarrow i}(h,w)\right),
\end{equation}
where attention is amplified at locations where the ego agent has high demand and the collaborator is reliable, and suppressed otherwise. The value features are further weighted by the perceptual certainty of the collaborator, $1 - U_j$, yielding the fused contribution $\tilde{\mathcal{V}}_{j \rightarrow i}$. The final enhanced BEV feature is obtained via residual aggregation across all collaborators:
\begin{equation}
    \hat{\mathcal{F}}_i = \mathcal{F}_i + \sum_{j \neq i} \tilde{\mathcal{V}}_{j\rightarrow i},
\end{equation}
which is subsequently passed to the perception head to produce the final collaborative result $\hat{\mathcal{O}}_i$.

\subsection{Training Strategy and Loss}

\textbf{Random masking.}
To prevent the ego agent from over-relying on collaborative information, a random masking strategy is adopted during training. Specifically, for each training sample, the messages from collaborating agents are randomly masked with probability $p_m$, forcing the ego agent to maintain strong single-agent perception capability while learning to exploit collaborative information when it is available.

\textbf{Perception loss.}
The perception loss $\mathcal{L}_{det}$ follows the standard CenterPoint detection head formulation, consisting of a focal loss for heatmap classification and a smooth-$\ell_1$ loss for box regression:
\begin{equation}
    \mathcal{L}_{det} = \mathcal{L}_{cls} + \lambda_{reg}\mathcal{L}_{reg},
\end{equation}
where $\lambda_{reg}$ is a balancing weight.

\textbf{Geometric uncertainty loss.}
To supervise the geometric uncertainty estimation, a KL-divergence-based loss is adopted following~\cite{CoAlign}. For center position and box dimensions, modeled as Gaussian distributions, the loss for each attribute $x$ is:
\begin{equation}
    \mathcal{L}_{x} = \frac{1}{2}\left(e^{-s_x}(\hat{x} - x_0)^2 + s_x\right),
\end{equation}
where $s_x = \log\sigma^2_x$ and $x_0$ is the ground truth. For the heading angle, modeled as a Von-Mises distribution, the loss is:
\begin{equation}
    \mathcal{L}_{\theta} = \log I_0(\kappa_\theta) - \kappa_\theta\cos(\hat{\theta} - \theta_0) + \lambda \cdot \mathrm{ELU}(s_\theta - c),
\end{equation}
where $I_0(\cdot)$ is the zero-order modified Bessel function and $\lambda$, $c$ are hyperparameters. The total geometric uncertainty loss is:
\begin{equation}
    \mathcal{L}_{geo} = \mathcal{L}_{x} + \mathcal{L}_{y} + \alpha_\theta\mathcal{L}_{\theta} + \alpha_s(\mathcal{L}_{w} + \mathcal{L}_{l}).
\end{equation}
The overall training objective is:
\begin{equation}
    \mathcal{L} = \mathcal{L}_{det} + \lambda_{geo}\mathcal{L}_{geo},
\end{equation}
where $\lambda_{geo}$ balances the perception loss and the geometric uncertainty supervision objective.
\begin{figure*}[t]
    \centering
    \begin{subfigure}[b]{0.48\linewidth}
        \centering
        \begin{minipage}[b]{0.48\linewidth}
            \includegraphics[width=\linewidth]{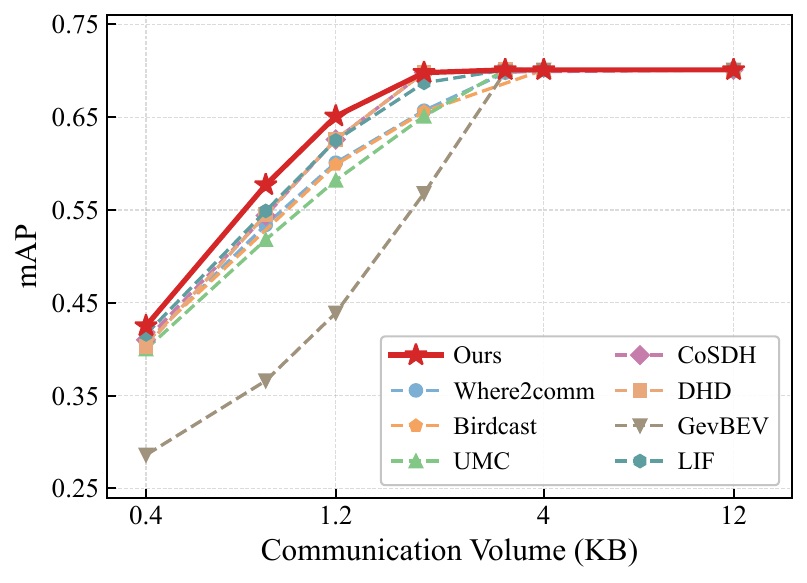}
        \end{minipage}
        \hfill
        \begin{minipage}[b]{0.48\linewidth}
            \includegraphics[width=\linewidth]{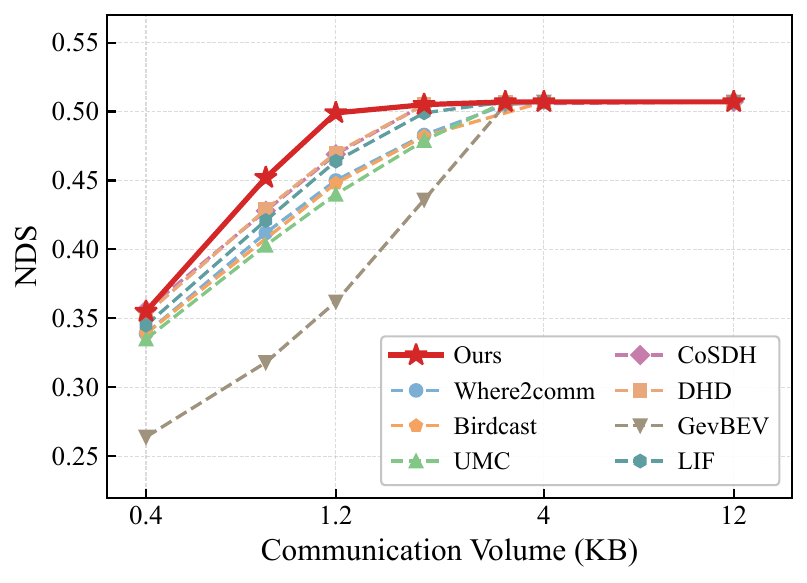}
        \end{minipage}
        \caption{Late-intermediate fusion paradigm.}
        \label{fig:late}
    \end{subfigure}
    \hfill
    \begin{subfigure}[b]{0.48\linewidth}
        \centering
        \begin{minipage}[b]{0.48\linewidth}
            \includegraphics[width=\linewidth]{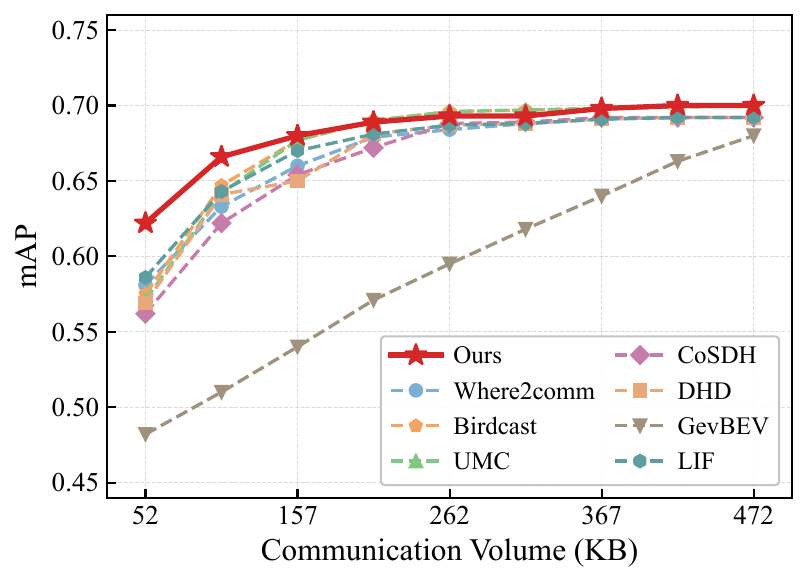}
        \end{minipage}
        \hfill
        \begin{minipage}[b]{0.48\linewidth}
            \includegraphics[width=\linewidth]{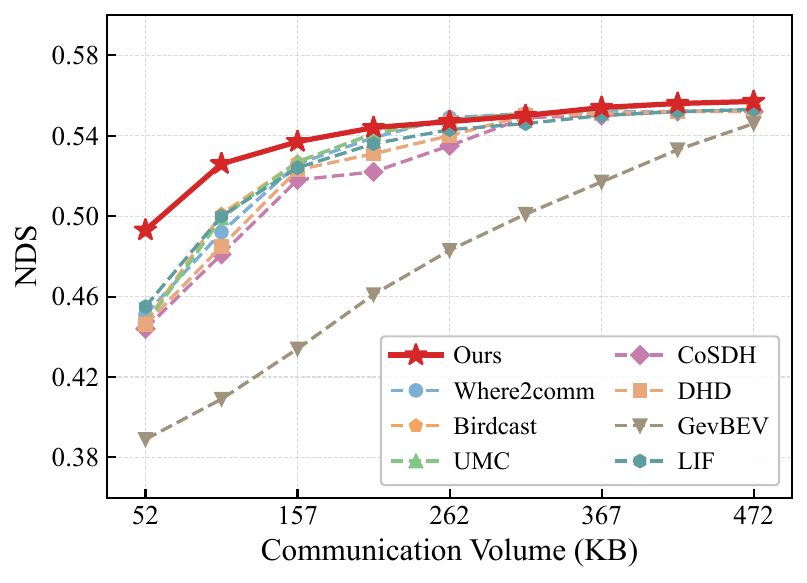}
        \end{minipage}
        \caption{Intermediate fusion paradigm.}
        \label{fig:inter}
    \end{subfigure}
    \caption{Evaluation on the UAV3D dataset. Ours {\modelname} achieves the best performance-bandwidth trade-off across both the late-intermediate paradigm (transmitting only perception outputs) and the intermediate paradigm (transmitting BEV feature maps).}
    \label{fig:figure04_1}
\end{figure*}

\begin{figure*}[t]
   \centering
   \begin{subfigure}[b]{0.48\linewidth}
       \centering
       \begin{minipage}[b]{0.48\linewidth}
           \includegraphics[width=\linewidth]{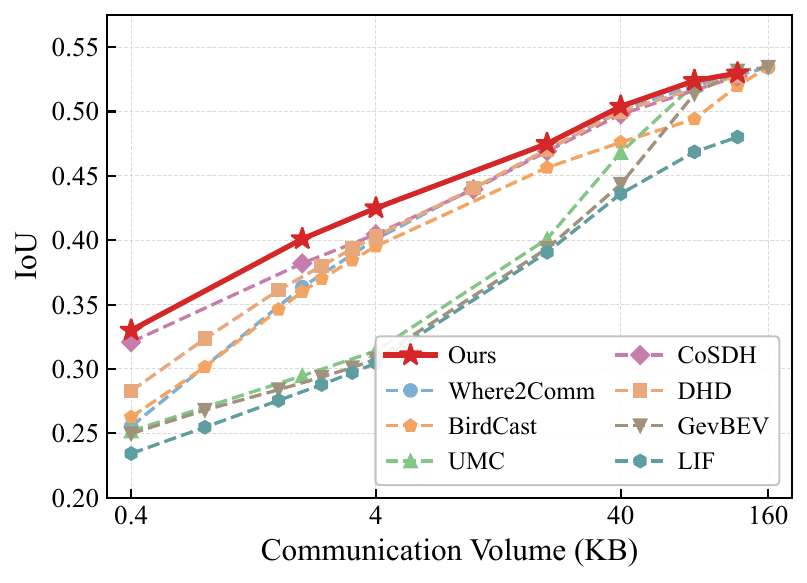}
       \end{minipage}
       \hfill
       \begin{minipage}[b]{0.48\linewidth}
           \includegraphics[width=\linewidth]{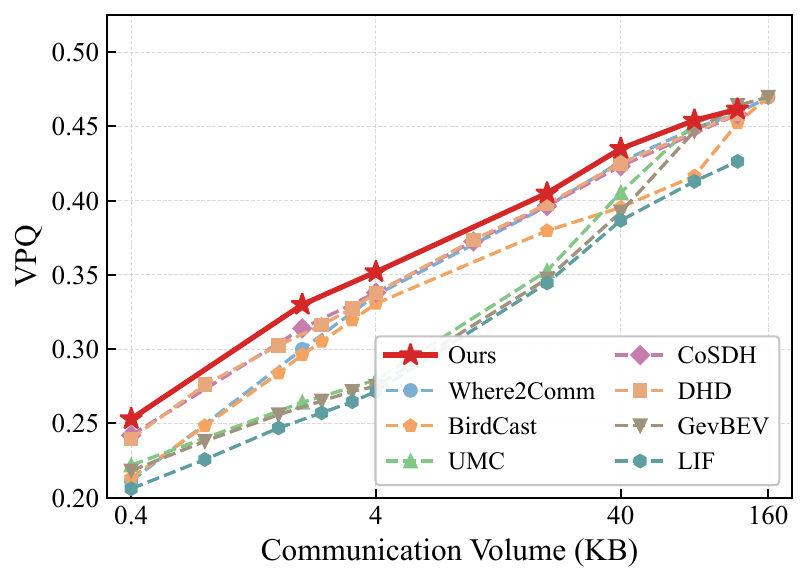}
       \end{minipage}
       \caption{Late-intermediate fusion paradigm.}
       \label{fig:dhd_late}
   \end{subfigure}
   \hfill
   \begin{subfigure}[b]{0.48\linewidth}
       \centering
       \begin{minipage}[b]{0.48\linewidth}
           \includegraphics[width=\linewidth]{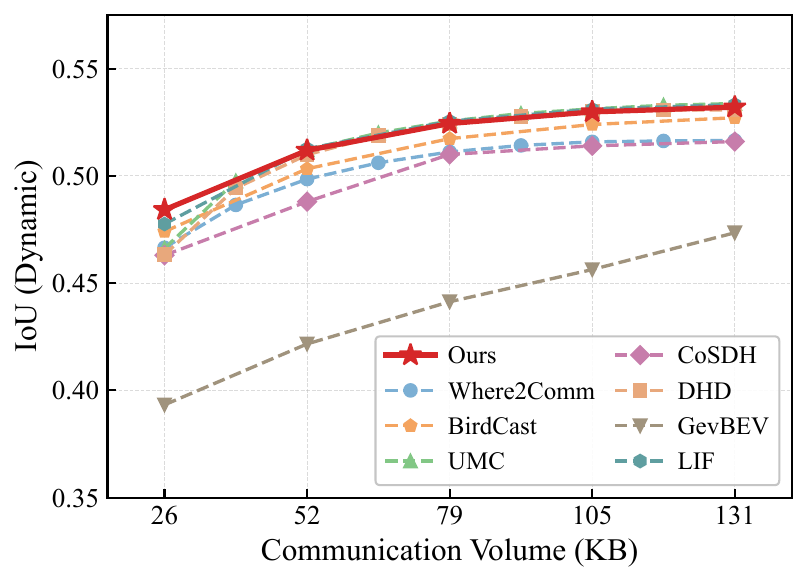}
       \end{minipage}
       \hfill
       \begin{minipage}[b]{0.48\linewidth}
           \includegraphics[width=\linewidth]{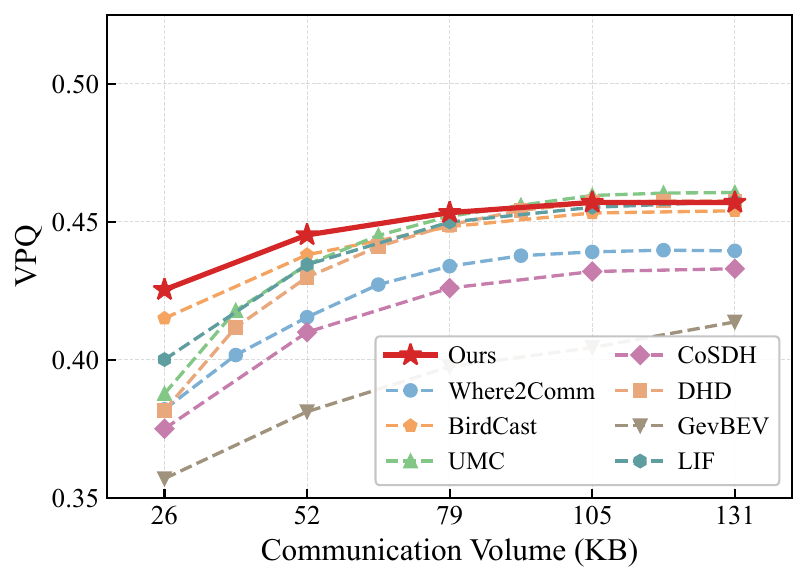}
       \end{minipage}
       \caption{Intermediate fusion paradigm.}
       \label{fig:dhd_inter}
   \end{subfigure}
   \caption{Evaluation on the Air-Co-Pred dataset. {\modelname} achieves the best performance-bandwidth trade-off across both the late-intermediate paradigm (transmitting only perception outputs) and the intermediate paradigm (transmitting BEV feature maps).}
   \label{fig:figure04_2}
\end{figure*}

\section{Experiments}

\subsection{Experimental settings}

\textbf{Datasets.}
We evaluate the proposed {\modelname} on three open-source UAV-based collaborative perception datasets, including two simulation datasets, UAV3D~\cite{UAV3D} and Air-Co-Pred~\cite{DHD}, and one real-world dataset, AGC-Drive~\cite{AGC-Drive}. In UAV3D, the perception area is set to $204.8 \times 204.8$ meters, where five UAVs act as collaborating agents and maintain a fixed formation at an altitude of approximately 60 meters.
Air-Co-Pred involves four UAVs acting as collaborating agents at traffic intersections. They operate at an altitude of approximately 50 meters, covering a ground area of about $100 \times 100$ meters, with images captured at a frequency of 2 Hz.
AGC-Drive involves one UAV and two equipped vehicles collecting synchronized data, with a perception range of $[-150, 150]$ meters along both the X and Y axes.

\textbf{Implementation.}
For the request and supply maps, the weighting hyperparameters are set to $\beta_1 = 0.7$, $\beta_2 = 0.3$, $\alpha_1 = 0.5$, and $\alpha_2 = 0.5$. For tasks that produce bounding box outputs, geometric uncertainty is additionally incorporated with $\alpha_3 = 0.1$. To supervise the geometric uncertainty estimation, we set $\alpha_{reg} = 2$, $\alpha_{center} = 0.25$, $\alpha_\theta = 0.125$, $\lambda_{elu} = 0.01$, $c = 1$, and $\lambda_{geo} = 0.2$. All models are trained on four NVIDIA RTX A6000 GPUs. 

\textbf{Evaluation metrics.}
For 3D object detection, we adopt average precision (AP) at intersection-over-union (IoU) thresholds of 0.5 and 0.7. For trajectory prediction, we employ IoU and Video Panoptic Quality (VPQ) for frame-level and video-level evaluation, respectively.

\subsection{Quantitative evaluation}
\textbf{Accuracy-bandwidth trade-Off comparison.}
We evaluate the performance of {\modelname} on the UAV3D and Air-Co-Pred datasets, focusing on the trade-off between accuracy and communication bandwidth. In both datasets, we compare two fusion paradigms: late-intermediate (transmitting only perception outputs) and intermediate (transmitting BEV feature maps). Our method achieves the best performance-bandwidth trade-off in both paradigms.

As shown in Fig.~\ref{fig:figure04_1}  and Fig.~\ref{fig:figure04_2}, {\modelname} outperforms other methods in terms of key metrics such as mAP, IoU, and VPQ while maintaining a relatively low communication volume. This demonstrates that {\modelname} strikes an optimal balance between high accuracy and efficient communication, making it an effective solution for real-world multi-UAV collaborative perception tasks.

\textbf{Comparison with SOTA performance.}
We compare {\modelname} with state-of-the-art methods on both the simulated Air-Co-Pred dataset and the real-world AGC-Drive dataset.

\begin{table}[t!]
\renewcommand{\arraystretch}{1.1}
\caption{Trajectory prediction results on the Air-Co-Pred dataset. \textcolor[RGB]{180,180,180}{DHD} denotes the official results, and $\ast$ denotes the results we reproduce.}
\centering
    \begin{adjustbox}{width=\linewidth}
    \begin{tabular}{c|c|c|c|c}
        \toprule
        \textbf{Method} & \textbf{Shared Data} & \textbf{Comm. (Mbps) $\downarrow$} & \textbf{IoU $\uparrow$} & \textbf{VPQ $\uparrow$} \\
        \midrule
        No Collaboration & / & 0 & 0.326 & 0.278 \\
        Early Collaboration & RGB images & $\sim$1000 & 0.545 & 0.459 \\
        Late Collaboration & perception outputs & $\sim$0.01 & 0.514 & 0.436 \\
        \cline{1-5}
        \addlinespace[2pt]
        V2X-ViT~\cite{V2X-ViT} & \multirow{9}{*}{feature maps} & \multirow{9}{*}{$\sim$100} & 0.533 & 0.457 \\
        V2VNet~\cite{V2VNet} & & & 0.538 & 0.469 \\
        Who2com~\cite{Who2com} & & & 0.443 & 0.370 \\
        When2com~\cite{When2com} & & & 0.458 & 0.404 \\
        Where2comm~\cite{Where2comm} & & & 0.514 & 0.442 \\
        UMC~\cite{UMC} & & & 0.523 & 0.443 \\
        UMC (w/o GRU) & & & 0.532 & 0.459 \\
        \textcolor[RGB]{180,180,180}{DHD~\cite{DHD}} & & & \textcolor[RGB]{180,180,180}{0.540} & \textcolor[RGB]{180,180,180}{0.462} \\
        {DHD$^\ast$} & & & 0.532 & 0.460 \\
        \cline{1-5}
        \addlinespace[2pt]
        \rowcolor{gray!20} {\modelname} (ours) & perception outputs & $\sim$0.01 & \textbf{0.535} & \textbf{0.470} \\
        \bottomrule
    \end{tabular}
    \end{adjustbox}
\label{table:DHD}
\vspace{-8pt}
\end{table}
\begin{table}[ht]
\centering
\caption{3D Detection Performance (\%) on AGC-Drive.}
\label{table:agcdrive}
\begin{tabular}{ccccc}
\toprule
Co-Mode & Model & mAP@0.5 & mAP@0.7\\
\midrule
\multirow{1}{*}{Late} & Lower-bound & 26.6 & 12.2 \\
\midrule
\multirow{1}{*}{Early}
& Upper-bound & 58.2 & 43.1 \\
\midrule
\multirow{4}{*}{Intermediate}
& V2VNet \cite{V2VNet} & 27.2 & 14.1 \\
& Cobevt \cite{CoBEVT} & 28.6 & 14.9 \\
& Where2comm \cite{Where2comm} & 34.8 & 22.5 \\
& V2X-ViT \cite{V2X-ViT} & 32.7 & 18.7 \\
\midrule
\multirow{2}{*}{Late-Intermediate}
& LIF \cite{LIF} & 34.5 & 22.7 \\
& \cellcolor{gray!10} UCP (Ours) & \cellcolor{gray!10} \textbf{36.2} & \cellcolor{gray!10} \textbf{23.7} \\

\bottomrule
\end{tabular}
\end{table}

Table~\ref{table:DHD} presents results on the Air-Co-Pred trajectory prediction benchmark. {\modelname} achieves the best performance with an IoU of 0.535 and a VPQ of 0.470, outperforming feature-map-based methods such as V2X-ViT~\cite{V2X-ViT} and Where2comm~\cite{Where2comm} while transmitting significantly less data, demonstrating the effectiveness of U-Fusion in aggregating compact semantic messages into informative collaborative representations. Table~\ref{table:agcdrive} further validates the generalizability of {\modelname} on the real-world AGC-Drive 3D object detection benchmark, where {\modelname} surpasses all competing methods with an AP@0.5 of 36.2\% and an AP@0.7 of 23.7\%, exceeding the strongest baseline LIF~\cite{LIF} by 1.7\% and 1.0\%, respectively.

\subsection{Qualitative Evaluation}
\begin{figure}[!t]
  \centering
  \includegraphics[width=\linewidth, keepaspectratio]{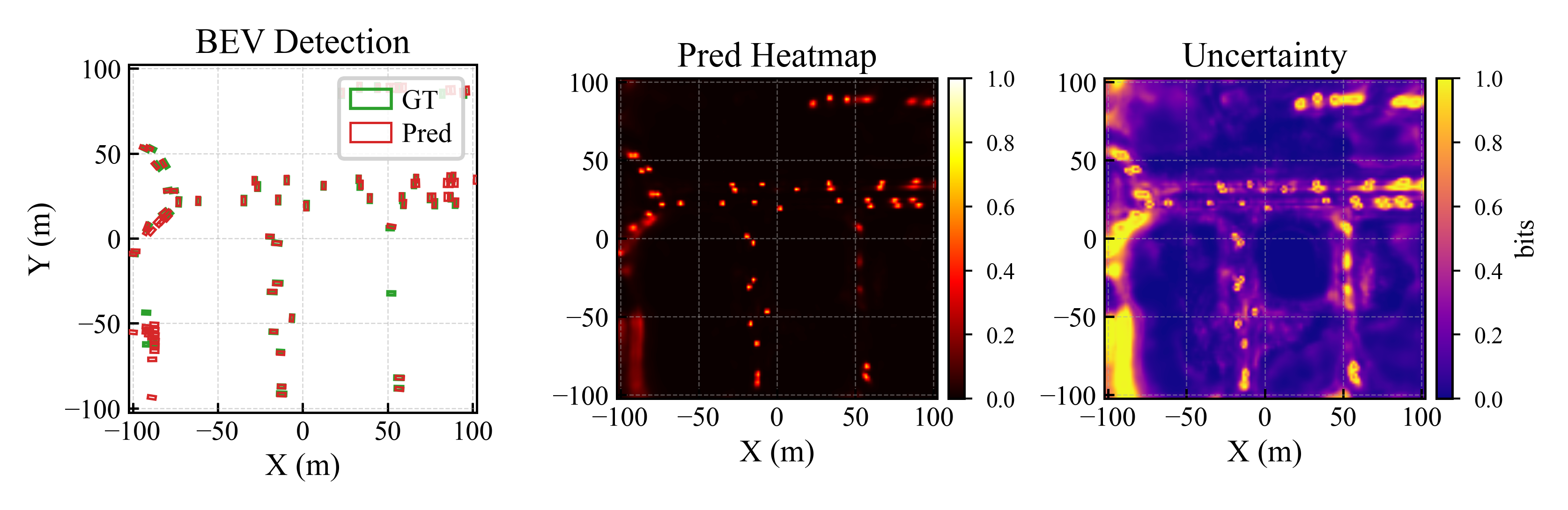}
  \caption{Visualization results of our {\modelname} on the UAV3D dataset.}
  \label{fig:vis}
\end{figure}

Fig.~\ref{fig:vis} presents the visualization results of the proposed method on the UAV3D dataset. The uncertainty map effectively identifies regions where the single-agent model suffers from either observation limitations or perceptually hard instances, and the spatial distribution of high-uncertainty regions closely aligns with the locations of missed and false detections in the BEV detection results. Prioritizing the exchange of complementary information in these regions enables the model to address both failure modes, thereby improving overall perception performance and robustness.

\begin{table}[]
\centering
\resizebox{0.49\textwidth}{!}{
\begin{tabular}{cc|cc|c|cc}
\hline
\multicolumn{2}{c|}{Request map} & \multicolumn{2}{c|}{Supply map} & \multirow{2}{*}{U-Fusion} & \multirow{2}{*}{IoU$\uparrow$} & \multirow{2}{*}{VPQ$\uparrow$} \\ \cline{1-4}
$U^{obs}$        &  $U^{cls}$        &  $Obj$               &  $Certainty$             &                              &                         &                         \\ \hline

$\checkmark$&&   &   &  \checkmark & 0.496 & 0.418 \\
$\checkmark$&$\checkmark$&&   & \checkmark  &0.501& 0.422\\
$\checkmark$&$\checkmark$&$\checkmark$& &   $\checkmark$ & 0.519 & 0.440 \\
$\checkmark$&$\checkmark$&$\checkmark$&$\checkmark$&$\checkmark$&0.522&0.443 \\ \hline
$\checkmark$&$\checkmark$&$\checkmark$&$\checkmark$&&0.513&0.431 \\ \hline
\end{tabular}
}
\caption{Ablation study of the key components in {\modelname} on the Air-Co-Pred dataset. $\mathrm{Obj}$ denotes the objectness map derived from the predicted classification scores, and $\mathrm{Certainty}$ denotes the perceptual certainty map defined as $1 - U_j$.}
\label{tab:ablation1}
\end{table}

\subsection{Ablation Studies}

Table~\ref{tab:ablation1} presents the ablation study on the Air-Co-Pred dataset. Incorporating classification uncertainty into the request map yields consistent gains over observational uncertainty alone, confirming that the two components capture complementary demand signals. On the supply side, adding the objectness map and certainty score progressively improves performance by prioritizing critical regions and suppressing unreliable transmissions. Removing the uncertainty-guided fusion module leads to a non-trivial performance drop, demonstrating that uncertainty-aware attention aggregation is essential for effective collaborative fusion.

\section{Conclusion}

In this work, we present \modelname, an uncertainty-driven collaborative perception framework for UAV-based multi-agent systems. By characterizing each spatial location through observational uncertainty and discriminative uncertainty, {\modelname} constructs spatial request and supply maps that enable efficient and targeted inter-agent communication. The received messages are further aggregated via U-Fusion, which conditions cross-attention on spatial demand and collaborator reliability. Extensive experiments on three benchmarks across 3D object detection and trajectory prediction demonstrate that {\modelname} achieves a superior performance-bandwidth trade-off while remaining heterogeneity-friendly and highly interpretable.

\addtolength{\textheight}{-12cm}

\bibliographystyle{IEEEbib}
\bibliography{references}

\end{document}